%% file: main.tex
\documentclass[final]{cvpr}

\usepackage{times}
\usepackage{epsfig}
\usepackage{graphicx}
\usepackage{amsmath}
\usepackage{amssymb}
\usepackage{xcolor}
\usepackage{caption}
\usepackage{stmaryrd}
\usepackage[pagebackref=true,breaklinks=true,colorlinks,bookmarks=false]{hyperref}

\def\cvprPaperID{1268} 
\def\confYear{CVPR 2021}
\definecolor{mygreen}{rgb}{0,.5,0.1}

\begin{document}

\title{Few-shot Image Generation via Cross-domain Correspondence}

\author{
Utkarsh Ojha$^{1,2}$\hspace{8mm}Yijun Li$^{1}$\hspace{8mm}Jingwan Lu$^{1}$\hspace{8mm}Alexei A. Efros$^{1,3}$\\
Yong Jae Lee$^{2}$\hspace{8mm}Eli Shechtman$^{1}$\hspace{8mm}Richard Zhang$^{1}$\vspace{2mm}\\ 
$^{1}$Adobe Research\hspace{15mm}$^{2}$UC Davis\hspace{15mm}$^{3}$UC Berkeley
}




\twocolumn[{%
\maketitle

\begin{center}
    \centering 
    \vspace{-0.18in}
    \includegraphics[width=0.99\textwidth]{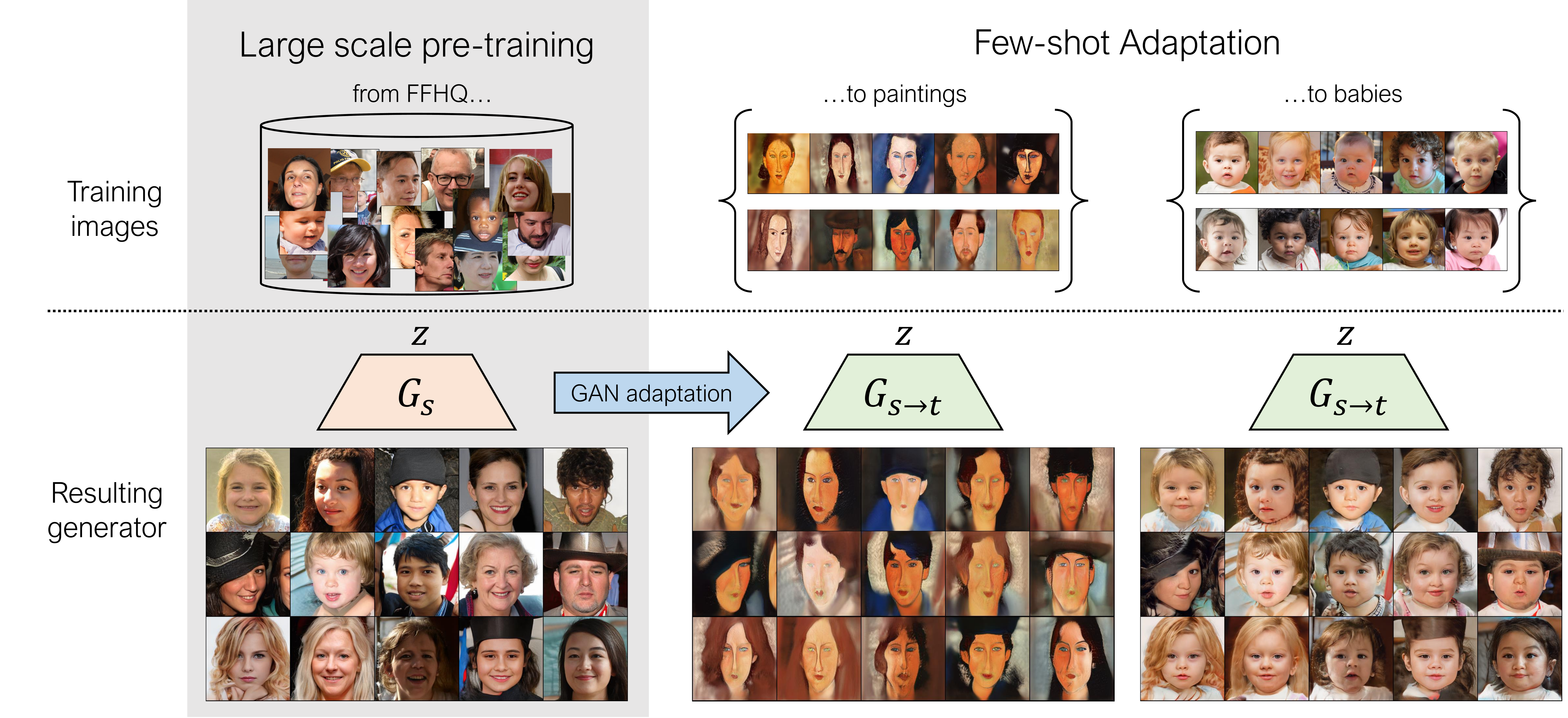}
    \vspace{-0.05in}
    \captionof{figure}{Given a model trained on a large source dataset ($G_s$), we propose to adapt it to arbitrary image domains, so that the resulting model ($G_{s\rightarrow t}$) captures these target distributions using extremely few training samples. In the process, our method discovers a one-to-one relation between the distributions, where noise vectors map to corresponding images in the source and target. Consequently, one can imagine how a natural face would look if Amedeo Modigliani had painted it, or how the face would look in its baby form. Please see our \href{https://utkarshojha.github.io/few-shot-gan-adaptation/}{webpage} for more results.}
    \label{fig:teaser}
\end{center}

}]


\begin{abstract}
Training generative models, such as GANs, on a target domain containing limited examples (e.g., 10) can easily result in overfitting.
In this work, we seek to utilize a large source domain for pretraining and transfer the diversity information from source to target.
We propose to preserve the relative similarities and differences between instances in the source via a novel cross-domain distance consistency loss.
To further reduce overfitting, we present an anchor-based strategy to encourage different levels of realism over different regions in the latent space.
With extensive results in both photorealistic and non-photorealistic domains, we demonstrate qualitatively and quantitatively that our few-shot model automatically discovers correspondences between source and target domains and generates more diverse and realistic images than previous methods.

\end{abstract}

\input{1_introduction}

\input{2_related}
\input{3_approach}

\input{4_results}

\vspace{-5pt}
\section{Conclusion and Limitations}

We proposed to adapt a pretrained GAN learned on a large source domain to a small target domain by discovering cross-domain correspondences. While our method generates compelling results, it is not without limitations. Cars $\rightarrow$ Abandoned cars in Fig.~\ref{fig:same_st} depicts an example, where the color of the red car changes to orange in its abandoned form, likely because of the existence of an orange car (and no red one) in the 10 training images. FFHQ $\rightarrow$ Sunglasses depicts another example, where a blonde hair turns dark with sunglasses. These show that there is a need for discovering better correspondence between the source and target domains, which will lead to more diverse generations. Nevertheless, we believe this work takes an important step towards creating more data-efficient generative models, demonstrating that existing source models can be leveraged in an effective way to model new distributions with less data.

\vspace{0.5em}
\noindent\textbf{Acknowledgements:} We thank Daichi Ito for the beautiful caricatures. Part of the work was supported through NSF CAREER IIS-1751206 and Adobe Data Science Research Award.  

{\small
\bibliographystyle{ieee_fullname}
\bibliography{egbib}
}

\clearpage
\input{supp}

\end{document}

%% file: 1_introduction.tex
\section{Introduction}

Consider 10 portrait paintings by the incomparable Amedeo Modigliani~\cite{yaniv2019artisticface}, shown in Fig.~\ref{fig:teaser} (middle). Given only these 10 paintings, would it be possible to train a model which can generate infinitely many paintings in the style of Modigliani?  Unfortunately, contemporary generative models~\cite{karras2020training, karras2019style, karras2019analyzing,vahdat2020NVAE,brock2018biggan} require thousands of images to train properly, not 10. This problem is of practical importance, since many such domains of interest have a very limited collection of images (e.g., there are just 10 examples per artist in the Artistic-Faces dataset~\cite{yaniv2019artisticface}).

Transfer learning serves as an alternative to training from scratch and has been explored in the context of generative adversarial networks (GANs) to address the limited data regime. The key idea is to start with a source model, pre-trained on a large dataset, and adapt it to a target domain with limited data by either making only small changes to the network parameters to preserve as much information as possible~\cite{wang2018transferring,noguchi2019image,wang2019minegan,mo2020freezeD,li2020fewshot}, or by synthetically increasing the training data via data augmentation~\cite{zhao2020diffaugment,karras2020training}. Most of these methods, however, are designed for scenarios with more than 100 training images. When the number of available images is lowered to just a few~\cite{li2020fewshot}, results often overfit to the training samples or are of poor quality. 

In this work, we explore transferring 
a different kind of information from the source domain, namely \emph{how images relate to each other}, to address the limited data setting. Intuitively, if the model can preserve the relative similarities and differences between instances in the source domain, then it has the chance to inherit the diversity in the source domain while adapting to the target domain.  To capture this notion, we introduce a novel cross-domain distance consistency loss, which enforces similarity in the distribution of pairwise distances of generated samples before and after adaptation. Unlike domain adaptation approaches like image-to-image translation, here we are adapting models, not images.  

Interesting properties emerge when enforcing this structure-level alignment between the two domains. Specifically, when the source and target domains are related (e.g., faces and caricatures), our approach automatically \emph{discovers} a one-to-one correspondence between them and is able to more faithfully model the true target distribution in terms of both diversity and image realism, as shown in Fig.~\ref{fig:teaser}. When the two domains are unrelated (e.g., cars and caricatures), our approach is unable to model the target distribution but still discovers interesting part-level correspondences to generate diverse samples.

Since the few training samples only form a small subset of the target distribution we seek to approximate, we find it necessary to enforce realism in two different ways, to not inordinately penalize the diversity among the generated images. We apply an image-level adversarial loss on the synthesized images which should map to one of the real samples. For all other synthesized images, we only enforce a patch-level adversarial loss. In this way, only a small subset of our generated samples need to look like one of the few-shot training images, while the rest are only forced to capture their patch-level texture.

\vspace{0.5em}
\noindent\textbf{Contributions.} Our main contribution is a novel GAN adaptation framework, which enforces cross-domain correspondence for few-shot image generation. Through extensive qualitative and quantitative results, we demonstrate that our model automatically discovers correspondences between related source and target domains to generate diverse and realistic images. 






%% file: 2_related.tex
\section{Related work}

\paragraph{Few-shot learning.} 
Representative few-shot classification approaches include learning a feature similarity function between the query and support examples~\cite{vinyals2016matching,snell2017prototypical} and learning to learn how to adapt a base-learner to a new task~\cite{finn2017MAML,nichol2018reptile}.

Few-shot image generation aims instead to hallucinate new and diverse examples while preventing overfitting to the few training images. Existing work mainly follows an adaptation pipeline, in which a base model is pretrained on a large source domain and then adapted to a smaller target domain. They either embed a small number of new parameters into the source model~\cite{noguchi2019image,wang2019minegan} or directly update the source model parameters, using different forms of regularization~\cite{mo2020freezeD,li2020fewshot}. Others employ data augmentation to reduce overfitting~\cite{zhao2020diffaugment,karras2020training} but are less effective under the extreme few-shot setting (e.g., 10 images).  In contrast to prior work, we regularize the adaptation of the source model by transferring how images relate to each other in the source domain to the target domain, leading to plausible generation results, even with very few examples.

\vspace{0.5em}
\noindent\textbf{Domain translation.} Translating images from the source domain is an alternative approach for generating more target domain data. However, such methods~\cite{isola2017image,Zhu-2017-cycleGAN,zhu2017bicyclegan} require a large amount of training data for both source and target domains and are not suitable for the few-shot scenario. Recent work~\cite{liu2019FUNIT,wang2020semiFUNIT,saito2020coco} has begun to address this issue via learning to separate the content and style factor, but requires large amount of labeled data (class or style labels). In our case, we assume access to a large amount of unlabeled data in the source domain and focus on adapting the source model to the target domain for unconditional image generation.

\vspace{0.5em}
\noindent\textbf{Distance preservation.} To alleviate mode collapse in GANs, DistanceGAN~\cite{benaim2017one} proposes to preserve the distances between input pairs in the corresponding generated output pairs. A similar scheme has been employed for both unconditional~\cite{tran2018dist,liu2019normalized} and conditional~\cite{mao2019mode,yang2019diversity} generation tasks to increase diversity in the generations. In our work, we aim to inherit the learned diversity from the source model to the target model and achieve this via our novel cross-domain distance consistency loss. 

%% file: 3_approach.tex
\section{Approach}



We are given a source generator $G_s$, trained on a large source dataset $\mathcal{D}_s$, which maps noise vectors $z\sim p_z(z)\subset \mathcal{Z}$, drawn from a simple distribution in a low-dimensional space, into images $x$. We aim to learn an adapted generator $G_{s\shortrightarrow t}$ by initializing the weights to the source generator and fitting it to a small target dataset $\mathcal{D}_t$. 



A naive translation 
can be obtained simply by using a GAN training procedure, with a learned discriminator $D$.
With the non-saturating GAN objective, this corresponds to solving: 
\begin{align}
\begin{split}
& \mathcal{L}_\text{adv}(G, D)=D(G(z))- D(x) \\
G_{s\shortrightarrow t}^{*} =& \mathbb{E}_{z\sim p_z(z), x \sim \mathcal{D}_t} \arg\min_G \max_D \mathcal{L}_\text{adv}(G, D).
\end{split}
\end{align}

\begin{figure}[t!]
    \centering
    \includegraphics[width=0.48\textwidth]{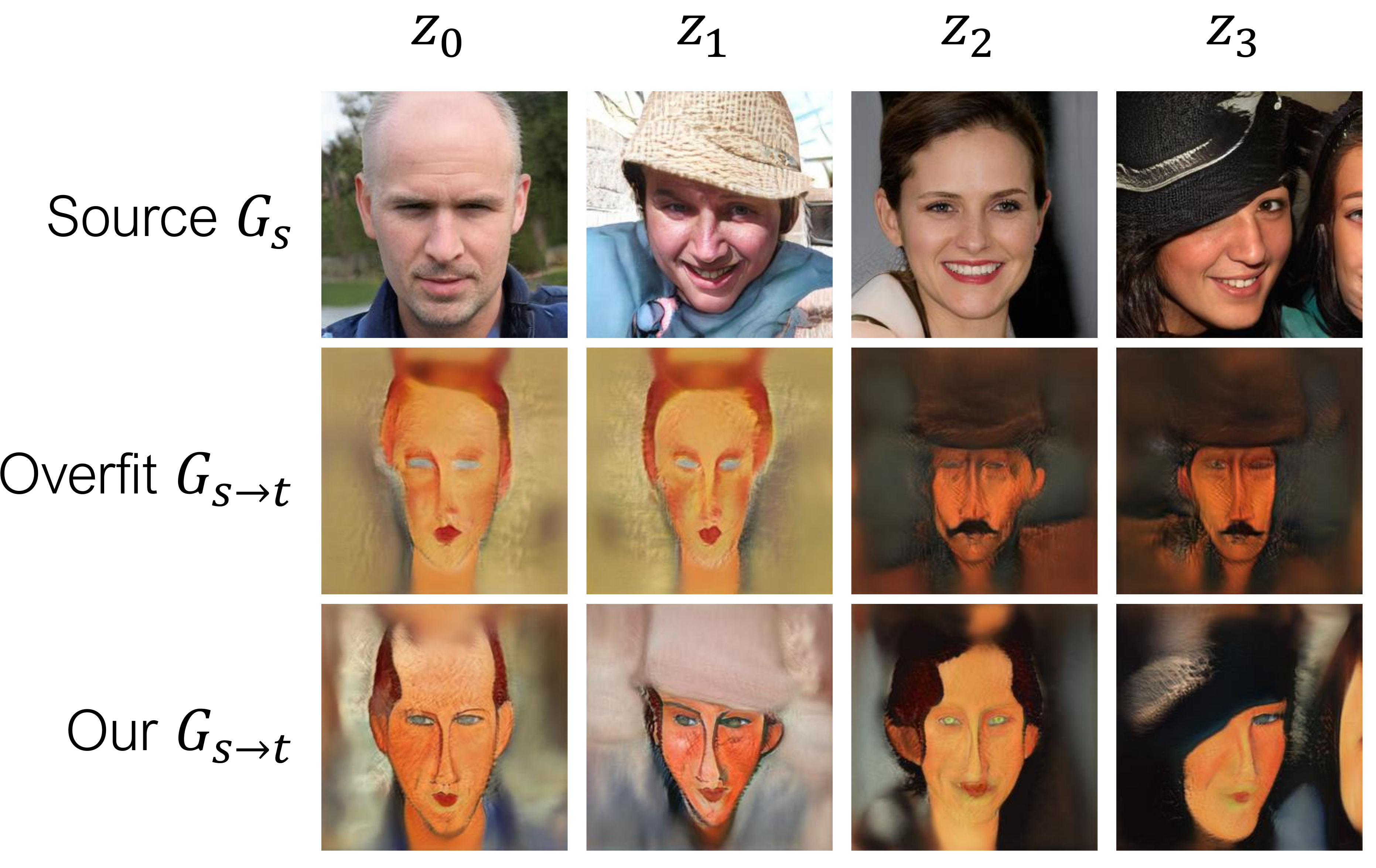}
    \caption{Adapting a source model (FFHQ) to 10 paintings using only the adversarial loss \cite{wang2018transferring} results in overfitting, which leads to a loss in correspondence between source and target images. Our adaptation method preserves this property in a much better way, where the same noise maps to corresponding images in source/target.} 
    \label{fig:overfit}
    \vspace{-1em}
\end{figure}

Previous work \cite{wang2018transferring} shows that this works well when the target dataset size exceeds 1000 training samples. However, in the
extreme few-shot setting, this method overfits, as the discriminator can memorize the few examples and force the generator to reproduce them.
This is shown in Fig.~\ref{fig:overfit}, where we see collapse after tuning the source model (top row) to the few-shot target dataset (middle row).

To prevent overfitting to generate diverse and realistic images (Fig.~\ref{fig:overfit}, bottom row), we propose a new cross-domain consistency loss (Sec.~\ref{sec:crossdomaindistanceconsistency}), which actively uses the original source generator to regularize the tuning process, and a ``relaxed'' discriminator (Sec.~\ref{sec:relaxedrealism}), which encourages different levels of realism over different regions in the latent space. Our approach is shown in Fig.~\ref{fig:approach}.

\subsection{Cross-domain distance consistency}
\label{sec:crossdomaindistanceconsistency}

A consequence of overfitting during adaptation is that relative distances in the source domain are not preserved. As seen in Fig.~\ref{fig:overfit}, the visual appearance between $z_1$ and $z_2$ collapses, disproportionately relative to $z_1$ and $z_3$, which remain perceptually distinct. We hypothesize that 
enforcing preservation of relative pairwise distances, before and after adaptation, will help prevent collapse.

\begin{figure}[t!]
    \centering
    \includegraphics[width=0.49\textwidth]{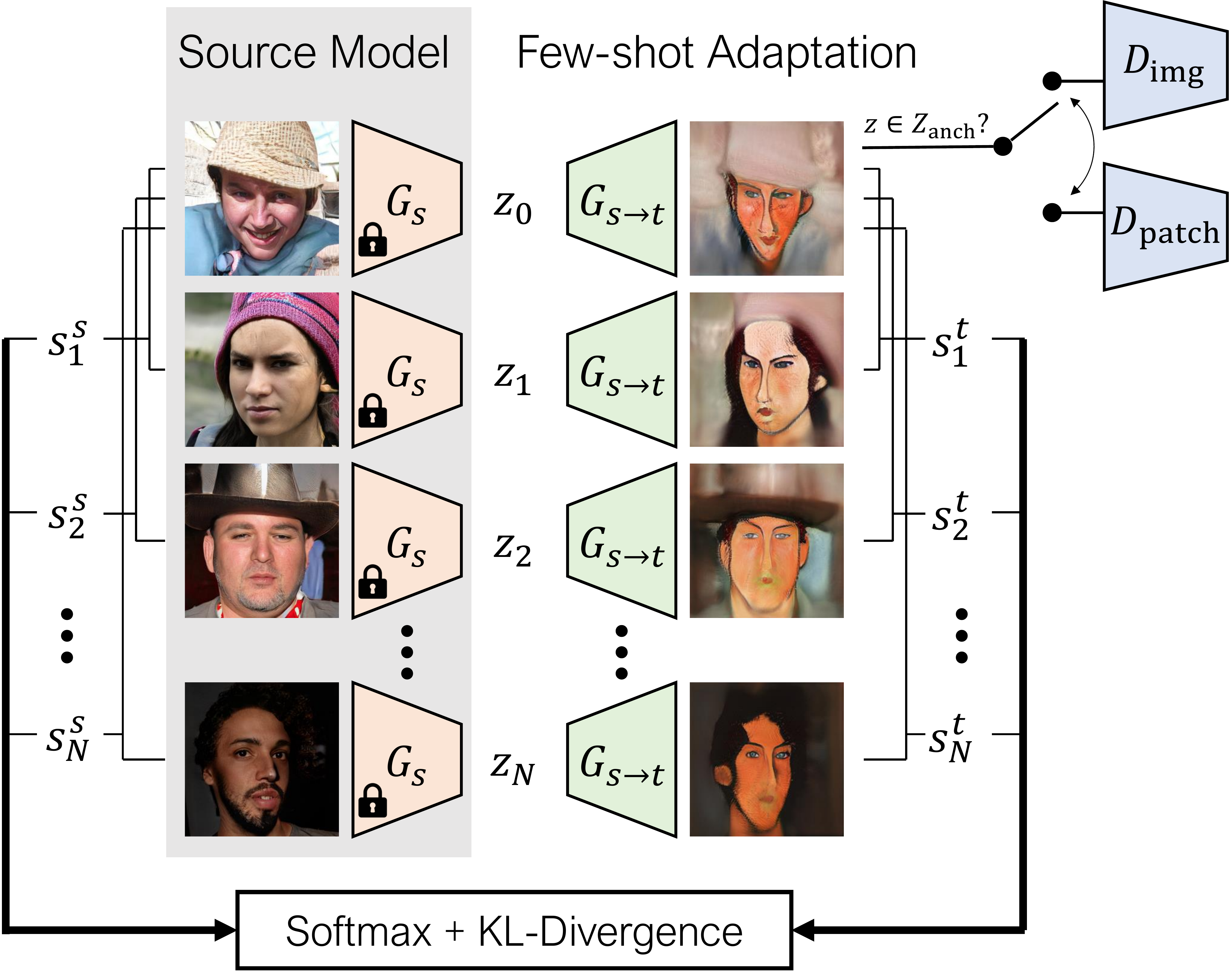}
    \caption{Our approach contains two key elements. (1) Cross-domain consistency loss $\mathcal{L}_\text{dist}$ aims to preserve the relative pairwise distances between source and target generations. In this case, the relative similarities between synthesized images from $z_0$ and other latent codes are encouraged to be similar. (2) Relaxed realism is implemented by using two discriminators, $D_\text{img}$ for noise sampled from the anchor region ($z_\text{anch}$) and $D_\text{patch}$ otherwise.
    } 
    \label{fig:approach}
    \vspace{-1em}
\end{figure}

To this end, we sample a batch of $N+1$ noise vectors $\{z_n\}_0^{N}$, and use their pairwise similarities in feature space to construct $N$-way probability distributions for each image. This is illustrated in Fig.~\ref{fig:approach} from the viewpoint of $z_0$. The probability distribution for the $i^\text{th}$ noise vector, for the source and adapted generators is given by,
\begin{equation}
\begin{split}
y^{s,l}_i = \text{Softmax}\big(\{\text{sim}(G_s^{l}(z_i), & G_s^{l}(z_j))\}_{\forall i\neq j}\big) \\
y^{s\shortrightarrow t,l}_i = \text{Softmax}\big(\{\text{sim}(G_{s\shortrightarrow t}^{l}(z_i)&, G^{l}_{s\shortrightarrow t}(z_j))\}_{\forall i\neq j}\big),
\end{split}
\vspace{-3pt}
\end{equation}
where $\text{sim}$ denotes the cosine similarity between generator activations at the $l^\text{th}$ layer. We are inspired by recent methods in contrastive learning~\cite{oord2018representation,chen2020simCLR,he2020momentum}, which converts similarities into probability distributions for unsupervised representation learning, as well as perceptual feature losses~\cite{johnson2016perceptual,dosovitskiy2016inverting,ulyanov2016texture}, which show that activations on multiple layers on discriminative networks help preserve similarity. We encourage the adapted model to have similar distributions to the source, across layers and image instances by using KL-divergence:
\begin{equation}
    \mathcal{L}_\text{dist}(G_{s\shortrightarrow t}, G_s) = \mathbb{E}_{\{z_i\sim p_z(z) \}} \sum_{l,i} D_{KL} \big( y_i^{s\shortrightarrow t,l} || y_i^{s,l}\big) .
    \vspace{-3pt}
\end{equation}
While correspondence helps prevent collapse, we also modify the adversarial loss to further prevent overfitting. 



\subsection{Relaxed realism with few examples}
\label{sec:relaxedrealism}

With a very small target data size, the definition of what constitutes a ``realistic'' sample becomes increasingly overconstrained, as the discriminator can simply memorize the few-shot target training set.
We note that the few training images only form a small subset of the desired distribution and extend this notion to the latent space. We define ``anchor'' regions, $\mathcal{Z}_\text{anch}\subset \mathcal{Z}$, which form a subset of the entire latent space.  When sampled from these regions, we use a full image discriminator $D_\text{img}$. Outside of them, we enforce adversarial loss using a patch-level discriminator $D_\text{patch}$,
\begin{align}
\begin{split}
\mathcal{L}'_\text{adv}(G, D_\text{img}, D_\text{patch}) = \mathbb{E}_{x \sim \mathcal{D}_t} & \big[ 
\mathbb{E}_{z\sim Z_\text{anch}} \mathcal{L}_\text{adv}(G, D_\text{img}) \\
+ & \mathbb{E}_{z\sim p_z(z)} \mathcal{L}_\text{adv}(G, D_\text{patch})
\big].
\vspace{-10pt}
\end{split}
\end{align}

To define the anchor space, we select $k$ random points, corresponding to the number of training images, and save them. We sample from these fixed points, with a small added Gaussian noise ($\sigma=.05$). We use shared weights between the two discriminators by defining $D_\text{patch}$ as a subset of the larger $D_\text{img}$ network~\cite{isola2017image, Zhu-2017-cycleGAN}; using internal activations correspond to patches on the input. The size depends on the network architecture and layer. We read off a set of layers, with effective patch size ranging from $22\times 22$ to $61\times 61$.





\subsection{Final Objective}

Our final objective consists of just these two terms: $\mathcal{L'}_\text{adv}$ for the appearance of the target and $\mathcal{L}_\text{dist}$, which directly leverages the source model to preserve structural diversity:
\begin{equation}
\begin{split}
G_{s\shortrightarrow t}^{*} = \arg\min_G \max_{D_\text{img}, D_\text{patch}} & \mathcal{L}'_\text{adv}(G, D_\text{img}, D_\text{patch}) \\ + \lambda \hspace{.5mm} & \mathcal{L}_\text{dist}(G, G_s).
\end{split}
\end{equation}

The patch discriminator gives the generator some additional freedom on the structure of the image. The adapted generator is directly incentivized to borrow the domain structure from the source generator, due to the cross-domain consistency loss. As shown in the top and bottom rows in Fig.~\ref{fig:overfit}, the model indeed discovers \textit{cross-domain correspondences} between source and target domains.

We use the StyleGANv2 architecture\footnote{\url{https://github.com/rosinality/stylegan2-pytorch}} \cite{karras2019analyzing}, pre-trained on a large dataset (e.g.~FFHQ \cite{karras2019style}) as our source model. We use a batch size of 4. Empirically, we find that a high $\lambda$, from $10^3$ to $10^4$, to work well. Additional training details can be found in the supplementary.

%% file: 4_results.tex
\begin{figure*}[t]
    \centering
    \includegraphics[width=1.\textwidth]{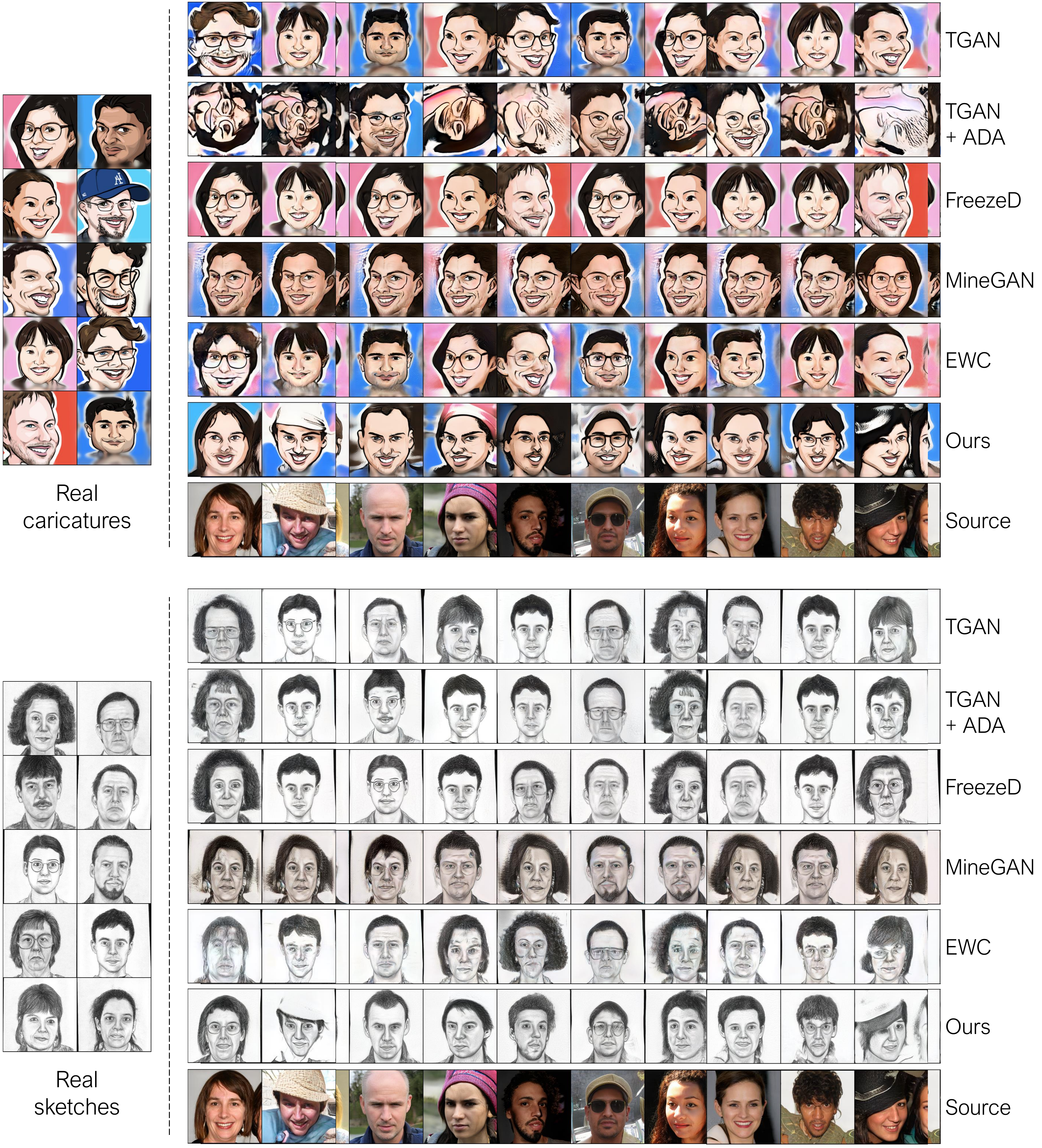}
    \caption{10-shot image synthesis results for different methods, which start from the same source model (bottom). Keeping the noise vectors same (across columns), we observe that the baselines either overfit, or only capture a few modes in the target domain. Our method generates higher quality and more diverse results which better correspond to the source domain images generated from the same noise.}
    \label{fig:img_gen}
    \vspace{-1.5em}
\end{figure*}

\section{Experiments}
We explore different $\texttt{source} \rightarrow \texttt{target}$ adaptation settings to analyze the effectiveness of our approach in preserving part-level correspondences between images generated from $G_s$ and $G_{s\rightarrow t}$. 
We also investigate what kinds of correspondences emerge when the source and target domains are unrelated.

\begin{figure*}[t]
    \centering
    \includegraphics[width=1.\textwidth]{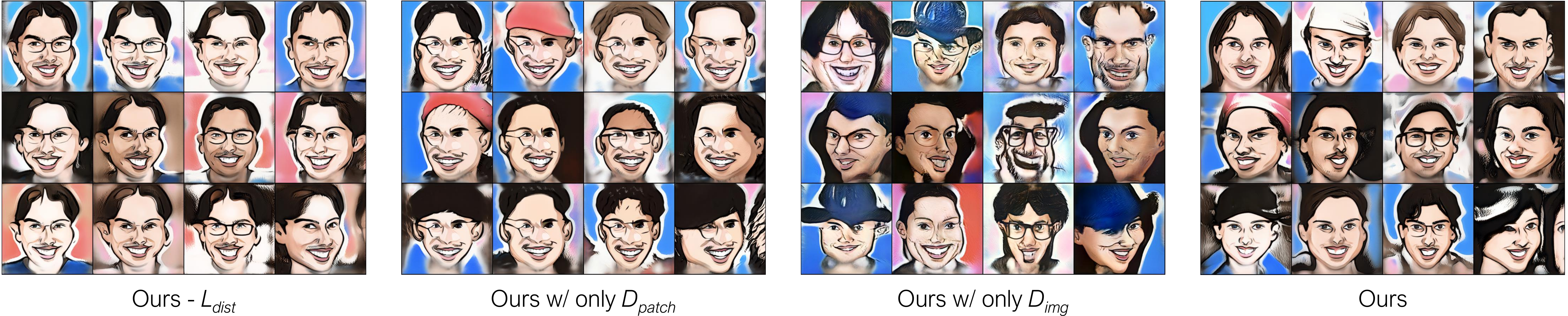}
    \caption{Effect of different components of our method. Absence of $\mathcal{L}_{dist}$ makes some properties of the images (e.g. hairline) look very similar. Application of only one of $D_\text{img}$ and $D_\text{patch}$ degrades the image quality by distorting the face structure.}
    \label{fig:ablation}
\end{figure*}

\begin{figure*}[t]
    \centering
    \includegraphics[width=1.\textwidth]{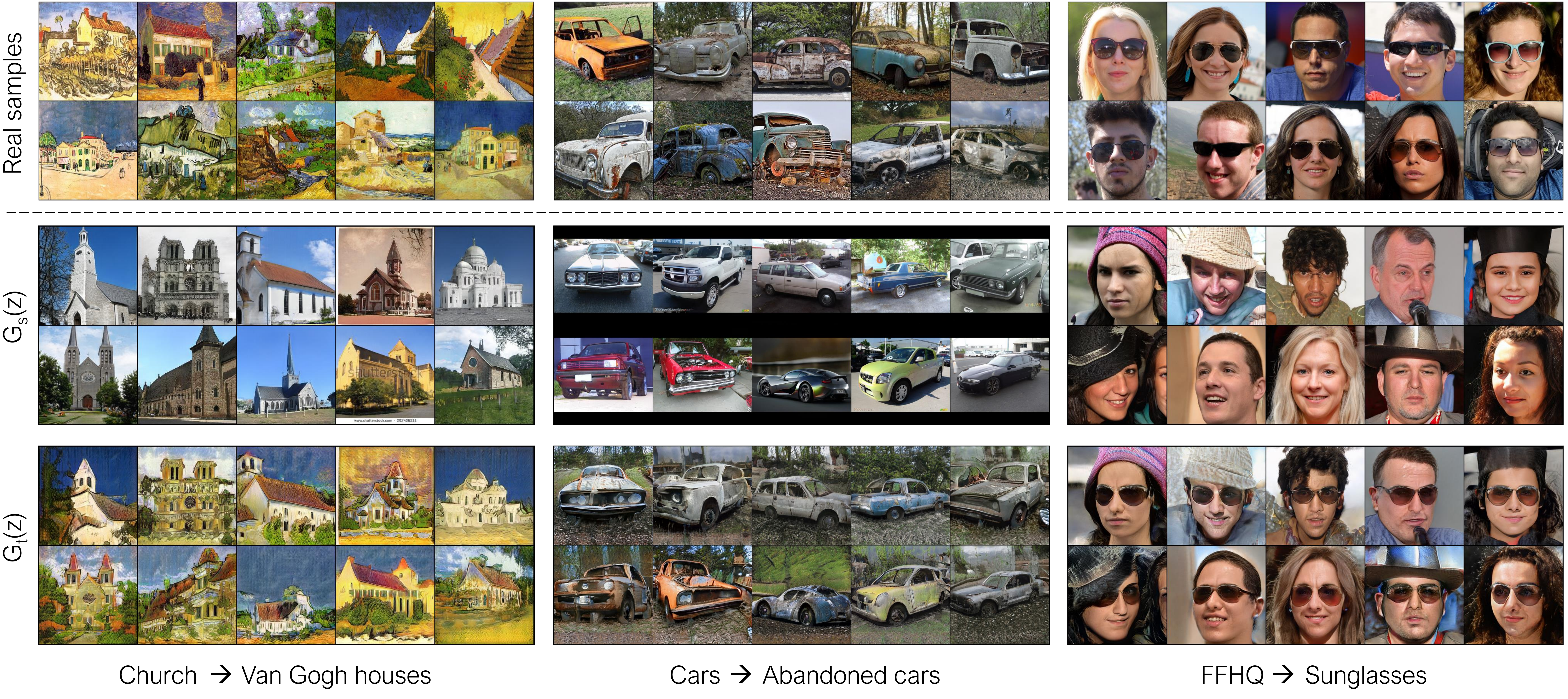}
    \caption{Visualizing the emerging correspondence in different adaptation settings. The generations from the domain of Van Gogh houses resemble the building structure from the source Church images. The generated wrecked/abandoned cars preserve the source car's body and pose. Generations from FFHQ $\rightarrow$ Sunglasses learn to add sunglasses to people's faces.}
    \label{fig:same_st}
    \vspace{-1em}
\end{figure*}


\vspace{0.5em}
\noindent\textbf{Baselines:} We compare to baselines, which similar to ours, adapt a pre-trained source model to a target domain with limited data. (i) Transferring GANs (TGAN) \cite{wang2018transferring}: finetunes a pre-trained source model to a target domain with the same objective used to train the source model; (ii) Batch Statistics Adaptation (BSA) \cite{noguchi2019image}: only adapts the scale and shift parameters of the model's intermediate layers; (iii) MineGAN \cite{wang2019minegan}: for a given pre-trained source (e.g. MNIST 0-8) and target (e.g. 9) domain, it transforms the original latent space of source to a space more relevant for the target (e.g. mapping all 0-8 regions to 4, being more similar to 9); (iv) Freeze-D \cite{mo2020freezeD}: freezes the high resolution discriminator layers during adaptation; (v) Non-leaking data augmentations \cite{karras2020training, zhao2020diffaugment}: uses adaptive data augmentations (TGAN + ADA) in a way that does not leak into the generated results; (vi) EWC \cite{li2020fewshot}: extends the idea of Elastic Weight Consolidation \cite{kirkpatrick2017EWC} for adapting a source model to a target domain, by penalizing large changes to \emph{important} weights (estimated via Fisher information) in the source model. 

\vspace{0.5em}
\noindent\textbf{Datasets:} We use source models trained on five different datasets: (i) Flickr-Faces-HQ (FFHQ) \cite{karras2019style}, (ii) LSUN Church, (iii) LSUN Cars, and (iv) LSUN Horses \cite{yu15lsun}. We explore adaptation to the following target domains: (i) face caricatures, (ii) face sketches \cite{wang2009sketches}, (iii) face paintings by Amedeo Modigliani \cite{yaniv2019artisticface}, (iv) FFHQ-babies, (v) FFHQ-sunglasses, (vi) landscape drawings, (vii) haunted houses, (viii) Van Gogh's house paintings, (ix) wrecked/abandoned cars. We operate on $256 \times 256$ resolution images for both the source and target domains. Adaptation is done on 10 images from the target domain, unless stated otherwise. 

\subsection{Quality and Diversity Evaluation}\label{sec:qual_eval}
\vspace{0.5em}
\noindent\textbf{Qualitative comparison} Fig.~\ref{fig:img_gen} shows results on two target domains using different methods, all of which start from the same source model pre-trained on FFHQ (Source).
We observe that TGAN strongly overfits to the available training data, as was the case for Amedeo paintings (Fig.~\ref{fig:overfit}). Using adaptive data augmentations (TGAN + ADA) has little to no effect on sketches, and further degrades the quality for caricatures, where augmentations (e.g. $90^{\circ}$ rotations) leak into the generated images. 
FreezeD, MineGAN and EWC perform better than TGAN by generating slightly more diverse images. However, the diversity is only introduced through minor modifications among the few captured modes. For example, (i) the caricature results for EWC show multiple generations capturing the same person with different expression/hairlines; (ii) the results on sketches using MineGAN depict a person with similar attributes in multiple generations. 
Our method better captures the distribution of caricatures and sketches, and generates diverse and realistic images containing objects which do not appear in the training images (e.g. hats, in sketches). This is because our method is flexible enough to not penalize the generated images which do not adhere to the 10 training samples. 

\vspace{0.5em}
\noindent\textbf{Quantitative comparison} The original Sketches, FFHQ-babies, and FFHQ-sunglasses datasets roughly contain 300, 2500, and 2700 images, respectively. To simulate a few-shot setting, we randomly sampled 10 images from each dataset to train our model. For evaluation purposes, however, we can use the entire dataset to measure how well our generated images model the true distribution. 

Table~\ref{tab:fid} shows the FID \cite{heusel2017-fid} scores. Our method significantly outperforms all baselines for Babies and Sunglasses. For domains with limited data, however, the FID score would not reflect the overfitting problem.

Ideally, we wish to assess the number of visually distinct images an algorithm can generate. In the worst case, the algorithm will simply overfit to the original $k$ training images. To capture this, we first generate 1000 images and assign them to one of the $k$ training images, by using lowest LPIPS distance~\cite{zhang2018LPIPS}.
We then compute the average pairwise LPIPS distance within members of the same cluster and then average over the $k$ clusters. A method that reproduces the original images exactly will have a score of zero by this metric.
Table~\ref{tab:lpips} summarizes the distances for different baselines over three target domains. We see that our method consistently achieves higher average LPIPS distances, indicating more distinct images being generated. We also visualize the cluster centers and their members, to see if they are semantically meaningful. See supp.~for details. 

\begin{table}[]
\scriptsize
\centering
\begin{tabular}{r|c|c|c}
         & \textbf{Babies} & \textbf{Sunglasses} & \textbf{Sketches} \\ \hline
\textbf{TGAN} \cite{wang2018transferring}    &      104.79 $\pm$ 0.03 &   55.61 $\pm$ 0.04        &  53.41 $\pm$ 0.02       \\
\textbf{TGAN+ADA} \cite{karras2020training} &     102.58 $\pm$ 0.12  &     53.64 $\pm$ 0.08      &    66.99  $\pm$ 0.01    \\
\textbf{BSA}    \cite{noguchi2019image}  &       140.34 $\pm$ 0.01 &      76.12 $\pm$ 0.01     &    69.32  $\pm$ 0.02    \\
\textbf{FreezeD} \cite{mo2020freezeD} &     110.92 $\pm$ 0.02  &    51.29  $\pm$ 0.05      &    46.54   $\pm$ 0.01   \\
\textbf{MineGAN} \cite{wang2019minegan}  &      98.23 $\pm$ 0.03 &   68.91 $\pm$ 0.03       &     64.34   $\pm$ 0.02 \\
\textbf{EWC}    \cite{li2020fewshot}  &     87.41 $\pm$ 0.02  &   59.73 $\pm$ 0.04       &     71.25   $\pm$ 0.01  \\
\textbf{Ours}     &   \textbf{74.39 $\pm$ 0.03}     &     \textbf{42.13 $\pm$ 0.04}       &     \textbf{45.67 $\pm$ 0.02}     
\end{tabular}
\vspace{3pt}
\caption{FID scores ($\downarrow$) for domains with abundant data. Standard deviations are computed across 5 runs.}
\label{tab:fid}
\end{table}

\begin{table}[]
\scriptsize
\centering
\begin{tabular}{r|c|c|c}
         & \textbf{Caricatures} & \textbf{Amedeo's paintings} & \textbf{Sketches} \\ \hline
\textbf{TGAN} \cite{wang2018transferring}    &    0.39 $\pm$ 0.06   &     0.41 $\pm$ 0.03       &     0.39 $\pm$ 0.03     \\
\textbf{TGAN+ADA}\cite{karras2020training} &    0.50 $\pm$ 0.05   &     0.51 $\pm$  0.04       &     0.41 $\pm$ 0.05   \\
\textbf{BSA} \cite{noguchi2019image}     &     0.35 $\pm$ 0.01 &     0.39 $\pm$ 0.04      &    0.35 $\pm$ 0.01     \\
\textbf{FreezeD} \cite{mo2020freezeD}  &    0.37 $\pm$ 0.01   &    0.40 $\pm$ 0.03      &     0.39 $\pm$ 0.03    \\
\textbf{MineGAN} \cite{wang2019minegan} &  0.39 $\pm$ 0.07     &    0.42 $\pm$ 0.03        &      0.40 $\pm$ 0.05    \\
\textbf{EWC}  \cite{li2020fewshot}    &  0.47 $\pm$ 0.03      &      0.52 $\pm$ 0.03     &     0.42 $\pm$ 0.03     \\
\textbf{Ours}     &  \textbf{0.53 $\pm$ 0.01}  &  \textbf{0.60 $\pm$ 0.01}         &     \textbf{0.45 $\pm$ 0.02}    
\end{tabular}
\vspace{3pt}
\caption{Intra-cluster pairwise LPIPS distance ($\uparrow$). Standard deviation is computed across the $k$ clusters ($k$ = no. of training samples).}
\label{tab:lpips}
\end{table}

\begin{figure*}[t]
    \centering
    \includegraphics[width=1.\textwidth]{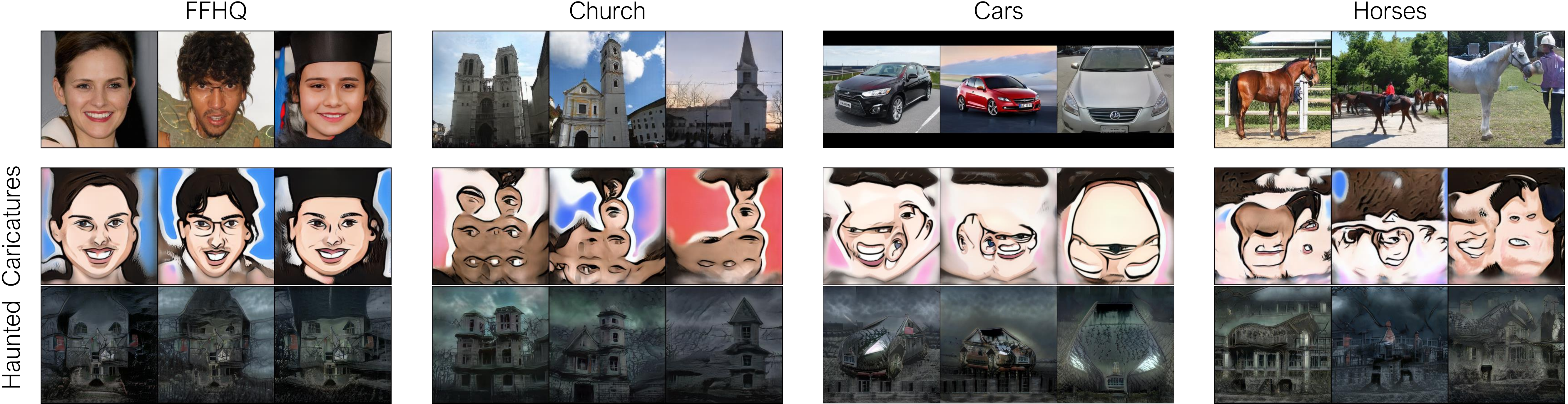}
    \caption{Visualizing the effect of unrelated source domains. In most cases, accurate modeling of a target distribution is not feasible when starting from an unrelated source domain. However, our method still discovers a correspondence on a part-level between the two domains, where different parts of the source (car's tires) correspond to parts of the target (caricature's eyes).}
    \label{fig:diff_st}
\end{figure*}

\vspace{0.5em}
\noindent\textbf{What role do different components of our method play?} 
We use caricature as the target domain, and first study the effect of our framework with and without $\mathcal{L}_\text{dist}$; see Fig.~\ref{fig:ablation}. We see that leaving out $\mathcal{L}_\text{dist}$ reduces diversity among the generations, all of which have very similar head structure and hair style. 
We next study the different ways we enforce realism. What happens if we keep $\mathcal{L}_\text{dist}$, but use image-level adversarial loss through $D_\text{img}$ on \emph{all} generations? `Ours w/ only $D_\text{img}$' results reveal the problem of mode collapse at the part level (same blue hat appears in multiple generations) and the phenomenon where some results are only slight modifications of the same mode (same girl with and without the blue hat). Could we then only use $D_\text{patch}$ to enforce patch-level realism on \emph{all} generations? `Ours w/ only $D_\text{patch}$' shows the results, where we observe more diversity, but poorer quality compared to `Ours w/ only $D_\text{img}$'. This is because the discriminator never gets to see a \emph{complete} caricature image, and consequently does not learn the part-level relations which makes a caricature look realistic. `Ours' combines all the ideas, resulting in generations which are diverse and realistic at both the part and image level.

\begin{figure}[t!]
    \centering
    \includegraphics[width=0.47\textwidth]{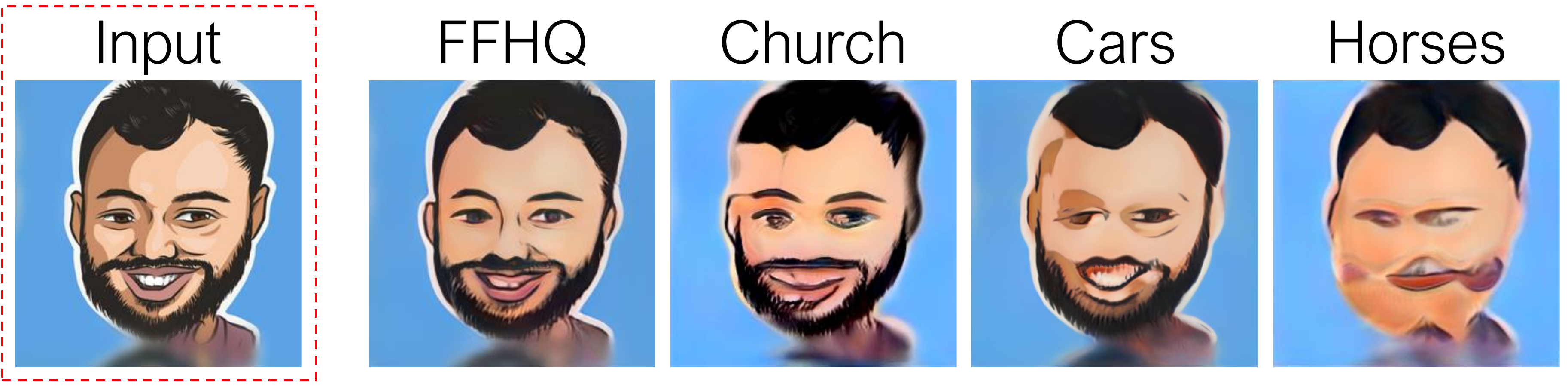}
    \caption{Embedding an unseen caricature image (left), and visualizing its reconstruction from models adapted from different source to the caricature domain.}
    \label{fig:recon}
    \vspace{-1em}
\end{figure}

\begin{figure*}[t!]
    \centering
    \includegraphics[width=1.\textwidth]{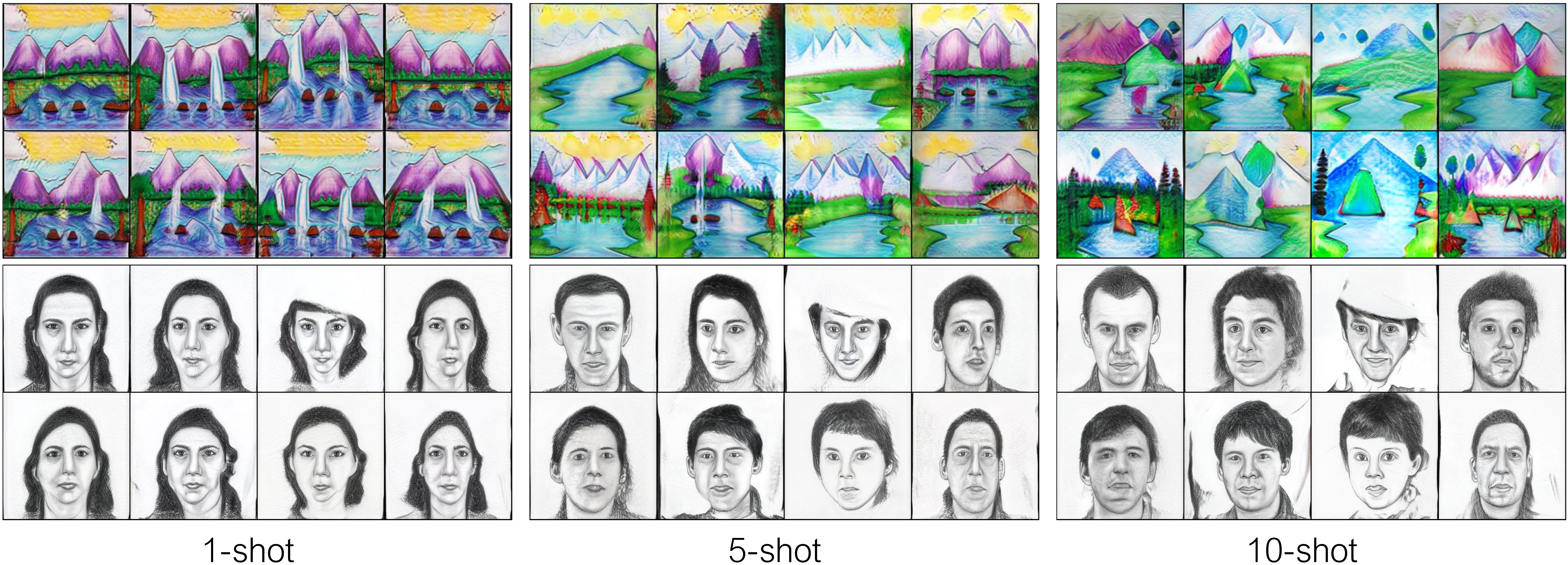}
    \caption{\textbf{Effect of training data size:} Even with just a single image, our method can capture it in different modes through the generations. Increasing the number of training samples results in more diversity in the generated sketches and landscapes.}
    \label{fig:size_abl}
    \vspace{-10pt}
\end{figure*}

\subsection{Analyzing source $\leftrightarrow$ target correspondence}

As mentioned before, our method captures, to a large extent, the correspondence between the images from source and target domains, and preserves it during the adaptation process. In this section, we study this property in detail in different source to target adaptation settings.

\vspace{0.5em}
\noindent\textbf{Related source/target domains} Our method builds around the idea of discovering correspondence between a source and a target domain. But how well does that property actually hold? We first consider the FFHQ $\rightarrow$ Caricatures/sketches setting in Fig.~\ref{fig:img_gen},and analyze the baselines' results. As seen previously in Sec.~\ref{sec:qual_eval}, the generated images mostly overfit to the training samples and are unable to borrow anything other than rough pose from the corresponding source. Using our method, on the other hand, we observe that the content of the natural source faces, as well as any of the accessories (e.g. hats/sunglasses), are preserved in both the sketch and caricature domain, depicting a much cleaner correspondence. We further study this for other source$\rightarrow$ target settings: (i) Church $\rightarrow$ Van Gogh houses, (ii) Cars $\rightarrow$ Abandoned cars, (iii) FFHQ $\rightarrow$ Sunglasses, as shown in Fig~\ref{fig:same_st}. 
When the source and target domains have similar semantics, the results generated from the same noise vectors in respective domains have clear correspondence. 

\vspace{0.5em}
\noindent\textbf{Unrelated source/target domains} We adapt four source models (FFHQ, Church, Cars, and Horses) to two target domains (Caricatures and Haunted houses) and present the results in Fig.~\ref{fig:diff_st}. For FFHQ $\rightarrow$ Caricatures and Church $\rightarrow$ Haunted, the generated results from the adapted model mimic the target domain appearance and reflect correspondences with the source. 
For all the remaining scenarios, the adaptations do not capture the target distribution accurately. However, some part-level correspondences still emerge in these settings. E.g., (i) Church $\rightarrow$ Caricatures: windows/doors of the Church roughly map to eyes of the caricature; (ii) Cars $\rightarrow$ Caricatures: wheels/bumpers of the cars adapt to represent eyes/mouth of caricatures respectively; (iii) Cars $\rightarrow$ Haunted: moon in the haunted houses (see supp. for the 10 images used) maps to the headlights, lighting them up; (iv) Horses $\rightarrow$ Haunted: bottom of the horse legs adapt to doors of haunted houses.  



\vspace{0.5em}
\noindent\textbf{Quantitative analysis of source/target relevance}
We translate four source models to four target domains: caricatures, haunted houses, landscape drawings, and abandoned cars. We then test how well a translated model can embed an \emph{unseen} image from the respective target domain, E.g., after translating the four source models to the caricature domain, we use a new caricature and embed it into the four models using Image2StyleGAN \cite{image2stylegan}. The results are shown in Fig.~\ref{fig:recon}. In Table~\ref{tab:unseen_recon}, for each domain, we report the average similarity scores between five unseen inputs and their reconstructions. We see that FFHQ, Church, and Cars best reconstruct images from caricatures, haunted houses, and abandoned cars respectively which aligns with our intuition.

\subsection{Effect of target dataset size}
So far, all results are generated with 10 training images per target domain. We now explore how the dataset size affects the quality and diversity of the generated images. We consider two adaptation setups, Church $\rightarrow$ Landscape drawings and FFHQ $\rightarrow$ Sketches, and present results in 1-shot, 5-shot and 10-shot settings; see Fig.~\ref{fig:size_abl}. The real images used in these settings can be found in the supplementary.  

In 1-shot, our method introduces small variations to the single target image, for example the lady appears in different poses in the generated sketches, and the mountains and waterfall have different structures. The diversity of the results increases with 5 training images. The sketches now reflect distinct identities. Further increasing the number of training samples (10-shot) introduces more details for the sketch domain, and generates more diverse landscapes.

\begin{table}[t!]
\footnotesize
\centering
\begin{tabular}{c|c|c|c}
               & \textbf{Caricature}  & \textbf{Haunted house} & \textbf{Wrecked car} \\ \hline
\textbf{FFHQ}    & \textbf{0.158 $\pm$ 0.045}       &  0.645 $\pm$ 0.024   &   0.643 $\pm$ 0.012     \\ 
\textbf{Church}     & 0.294 $\pm$ 0.077     &   \textbf{0.599 $\pm$ 0.028} &      0.621 $\pm$ 0.032 \\ 
\textbf{Cars}     & 0.233 $\pm$ 0.106       &   0.635 $\pm$ 0.031  &      \textbf{0.606 $\pm$ 0.057}  \\ 
\textbf{Horses} &  0.299 $\pm$ 0.083     &  0.631 $\pm$ 0.028   &      0.619 $\pm$ 0.038 
\end{tabular}
\vspace{-5pt}
\caption{Relevance of source and target domains, measured via LPIPS ($\downarrow$) between an unseen image and its reconstruction. Better the reconstruction $\rightarrow$ more similar domains.}
\label{tab:unseen_recon}
\vspace{-8pt}
\end{table}

%% file: supp.tex
\setcounter{section}{0}

\section*{Supplementary}
This document provides additional information about the proposed method. First, we continue the discussion on training and architecture details of our method and the baselines. We use the official pre-trained models for Church, Cars and Horses \cite{karras2019analyzing}, and a 256 x 256 resolution pre-trained model for FFHQ.\footnote{We use \url{https://github.com/rosinality/stylegan2-pytorch}}. The generator's ($G$) and the discriminator's ($D_{img}$) architecture are the same as StyleGAN2. For $D_{pch}$, we consider the first $l$ layers of $D_{img}$, and convert the corresponding feature to a $N$$\times$$N$ output, where each member's receptive field corresponds to a patch in the input image. We use Adam optimizer \cite{kingma2019adam}, and use the rest of the hyperparameters (e.g. learing rate) from \cite{karras2019analyzing}. The ideal training duration for adapting to different target domains is as follows: for domains whose appearance matches very closely with the source, i.e. babies/sunglasses with FFHQ as source, we get good results within 1000 iterations. For other face-based target domains (e.g. Modigliani's paintings), we train our models for 5000 iterations to get decent results. For more complex domains (e.g. landscape drawings), we observe our best results at around 10000 iterations. Note that in eq. 4 (main paper), we have used $z\sim p_z(z)$ instead of $z\sim {p_z(z) - Z_{anch}}$ for the second term. This is because if an image from $Z_{anch}$ is (globally) realistic, its patches will be realistic as well. Also note that the other way around is not true.

\paragraph{Details about MineGAN} We use the publicly available code of MineGAN\footnote{\url{https://github.com/yaxingwang/MineGAN/tree/master/styleGAN}} to produce its results (Fig. 4, Tab 1/2). However, since there is an involvement of an additional \emph{miner} network during the adaptation process, we tried experimenting with smaller networks (than the default) for the extreme few-shot setting (10 training images). The results obtained were similar to the default setting -- FID: 96.72, 68.67 for babies and sunglasses domain respectively, suggesting that reducing the complexity of the miner network alone is not sufficient to achieve better results.

\begin{figure*}[h]
    \centering
    \includegraphics[width=1.\textwidth]{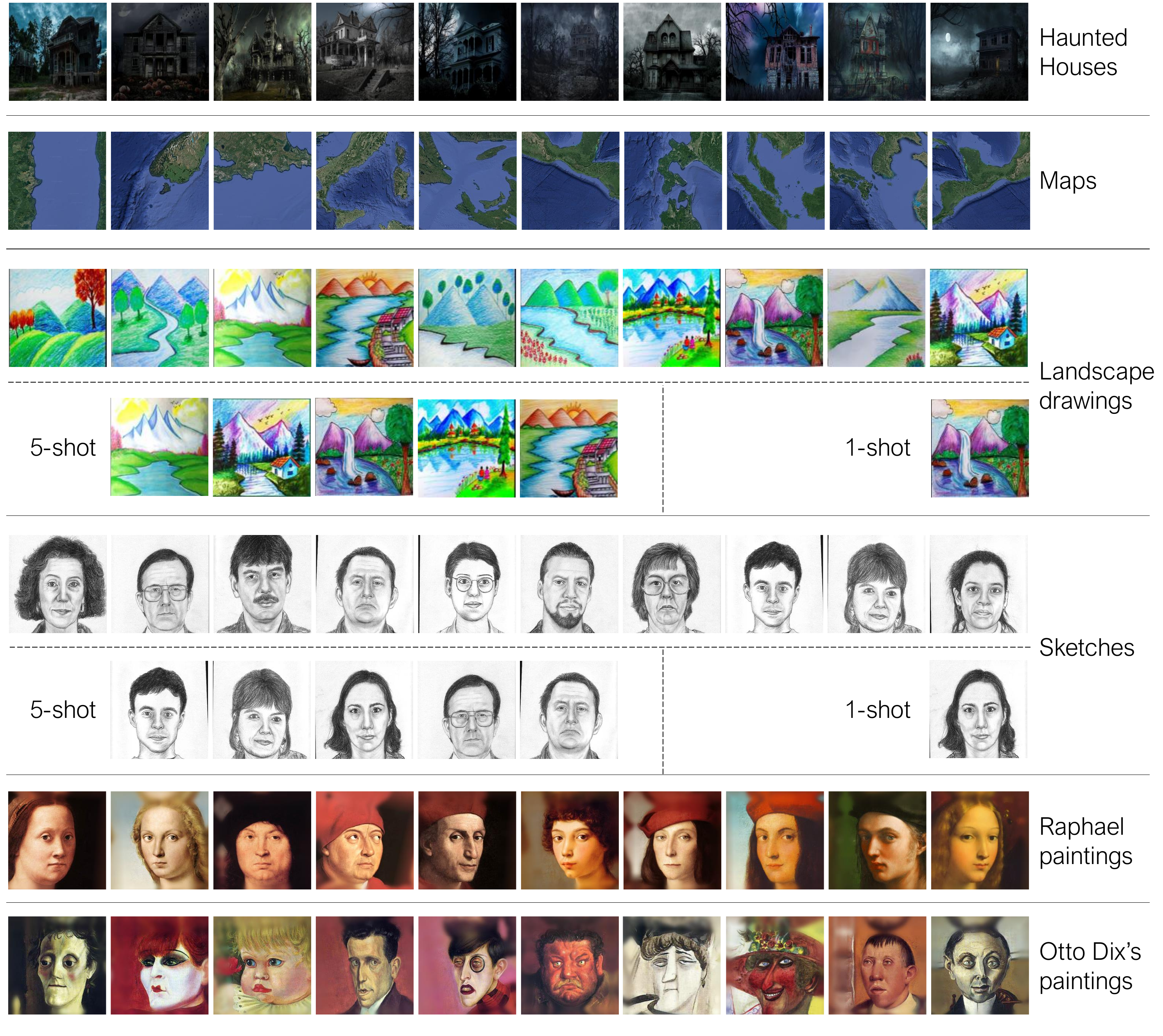}
    \caption{Real images used for translating a source model to different target domains. For landscape drawings and sketches, we have shown the images used in 1-shot and 5-shot scenario, for the experiment presented in Fig. 9 of the main paper, and Fig.~\ref{fig:size_ablation_full} of this document.}
    \label{fig:real_img}
    \vspace{-1.5em}
\end{figure*}

\paragraph{FFHQ $\rightarrow$ face domains} Fig.~\ref{fig:real_img} shows the real images used for different target domains in our experiments (apart from those presented in the main paper). Next, we show the results of translating a source model trained on natural faces (FFHQ) to different kinds of target domains. We observe the diversity in the generated images, which come as a result of preserving correspondence between the source and target distribution. Fig.~\ref{fig:embed_full} shows more examples for the idea discussed in Fig. 8 of the main paper. These four caricature images are unseen during the adaptation of a source model X (X is FFHQ/Church/Cars/Horses) to the caricature domain. We again observe that adapting FFHQ (natural faces) to the caricature domain best embeds and reconstructs unseen images, indicating that caricature as a domain is most related to FFHQ than any other source domain. Fig.~\ref{fig:size_ablation_full} presents an extension of Fig. 9 of main paper, where we study the 1-shot, 5-shot and 10-shot setting for two baselines, and compare it with our method. We notice that both FreezeD and EWC,  overfit to the target sample in 1-shot setting, generating virtually identical sketches/scenes. This trend of overfitting for these baselines continues in 5/10-shot settings as well, where generations collapse to small variations around a few modes. Our method, on the other hand, takes the benefit of increasing training data size, by learning to generate more and more diverse samples, different from the images used for training.

\paragraph{Visualizing the clusters} Fig.~\ref{fig:clusters} visualizes the clustering-based diversity assessment introduced in Sec. 4.1 of the main paper. We group the generated images from a method into $k$ clusters, with the $k$ training images serving as the cluster center. After this, we study the resulting clusters, where we visualize how similar is (i) the closest member to the center (measured via LPIPS), (ii) the farthest member to the center. The intuition is that a method whose generated images overfit to the training data will result in clusters where the closest member is very similar to the corresponding cluster center. Each column deals with one cluster, where the cluster centers (real images used for training) are shown in the middle. The top half of the figure visualizes the closest members for different methods, whereas the bottom half visualizes the farthest ones. When no images get assigned to a cluster, the concept of closest/farthest members doesn't apply, and we depict this with a red cross. Summarily, we observe that the closest members from TGAN/EWC are much more similar to the corresponding center than our method, whose even closest members are visually distinct. This observation also helps explain the better performance of our method compared to others in Table 2 (main paper).

\paragraph{Hand gestures experiment} We find the property of emerging correspondences within seemingly unrelated source/target domains interesting, and hence for creativity purposes, take a further step to explore the idea. We collect images of arbitrary hand gestures being performed over a plain surface, and train a \emph{source} model from scratch using that dataset. Next, we adapt it to various domains such as landscapes, fire, maps. During inference, we observe different aspects of the target domains a pair of hands can control (e.g. structure of river/islands). Please see our teaser video, which shows the correspondence results in this case, as well as better explains the benefits of our method in previously discussed scenarios (e.g. FFHQ $\rightarrow$ caricatures, Church $\rightarrow$ Van Gogh houses).

\paragraph{Precision and recall metrics} A limitation of FID~\cite{heusel2017-fid} is that it packs two aspects of the generated images, sample quality and diversity, into one score. This makes it difficult to disentangle and study the two properties separately. To overcome this, density and coverage metrics were proposed to evaluate the generative models \cite{ferjad2020icml}. In some feature space (e.g. CNN embeddings), density measures how many real-sample neighbourhood regions contain a fake sample. Coverage, in the same space, measures the ratio of real samples whose neighbourhood contains at least one fake sample. In both the definitions, \emph{neighbourhood} is defined as a spherical region around a real sample, with its radius given by the distance from the next nearest real sample. A high score for both the metrics is preferred. Density is unbounded, whereas coverage is bounded by 1. We present evaluation of the baselines fare using these metrics on FFHQ babies dataset in Table~\ref{tab:dc}. We observe that MineGAN achieves a superior \emph{density} score, i.e. quality of the generated image, but suffers in the \emph{coverage} aspect. This is again an indication of mode collapse to a small number of high quality samples. Our method achieves a better balance between the quality as well as diversity of the generated samples. Note that this result is in alignment with the one presented in Table 2 (main paper), which studies diversity among the generated samples in a different way.

\begin{table}[]
\scriptsize
\centering
\begin{tabular}{r|c|c}

         & Density & Coverage\\ \hline
\textbf{TGAN} \cite{wang2018transferring}    &      0.379 &   0.250     \\
\textbf{TGAN+ADA} \cite{karras2020training} &    0.434  &   0.285     \\
\textbf{FreezeD} \cite{mo2020freezeD} &     0.418  &    0.217       \\
\textbf{MineGAN} \cite{wang2019minegan}  &      \textbf{0.803} &  0.125  \\
\textbf{EWC}    \cite{li2020fewshot}  &     0.301  &   0.325   \\
\textbf{Ours}     &   0.690     &     \textbf{0.467}        
\end{tabular}
\vspace{3pt}
\caption{Density ($\uparrow$) and Coverage ($\uparrow$) scores for FFHQ babies.}
\label{tab:dc}
\end{table}


\begin{figure*}[t]
    \centering
    \includegraphics[width=1.\textwidth]{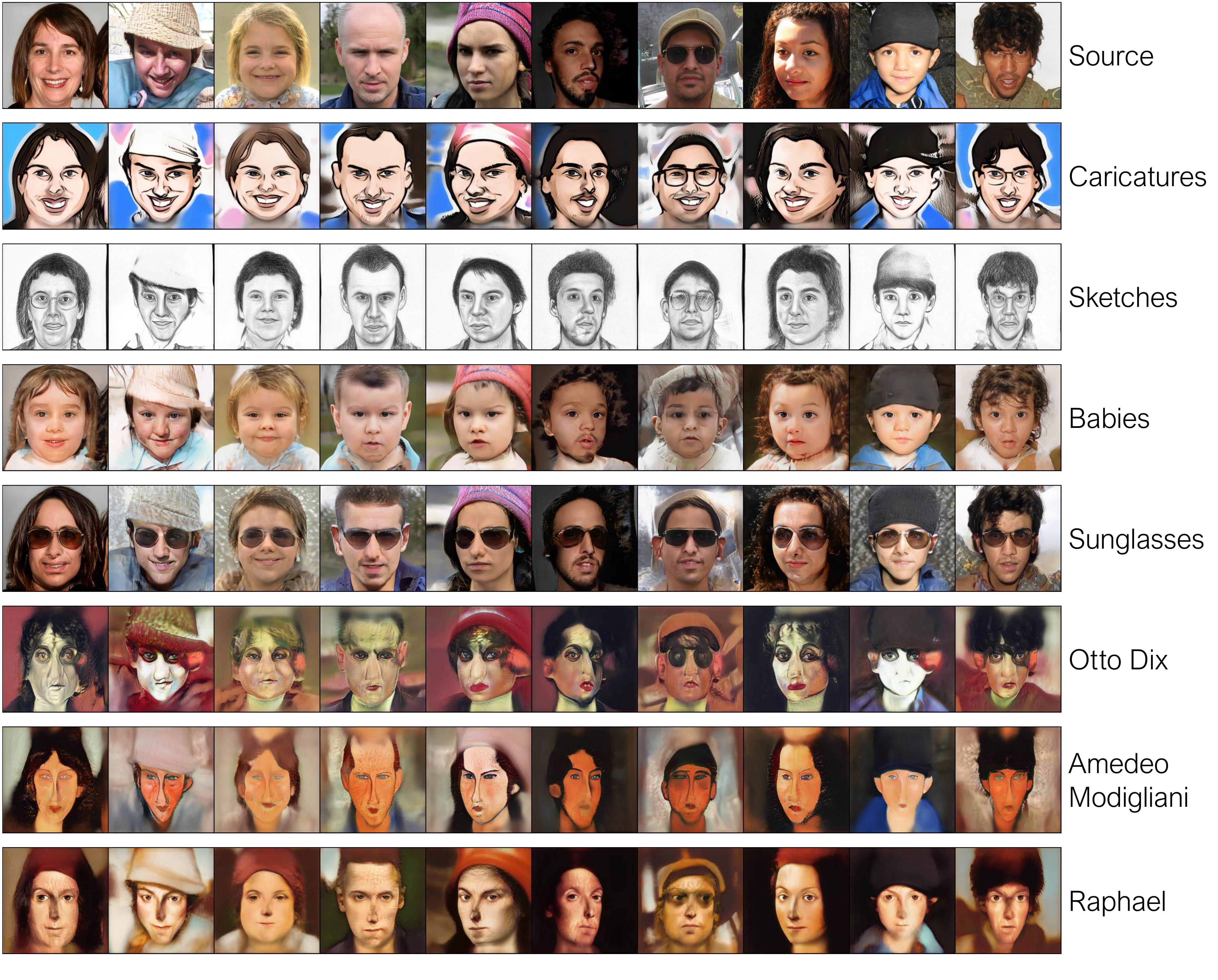}
    \caption{Translating the FFHQ source model to different target domains. The noise vector is kept same across the columns, so that we can study the relation between the corresponding source and target image.}
    \label{fig:face_domains}
    \vspace{-1.5em}
\end{figure*}

\begin{figure*}
    \centering
    \includegraphics[width=1.\textwidth]{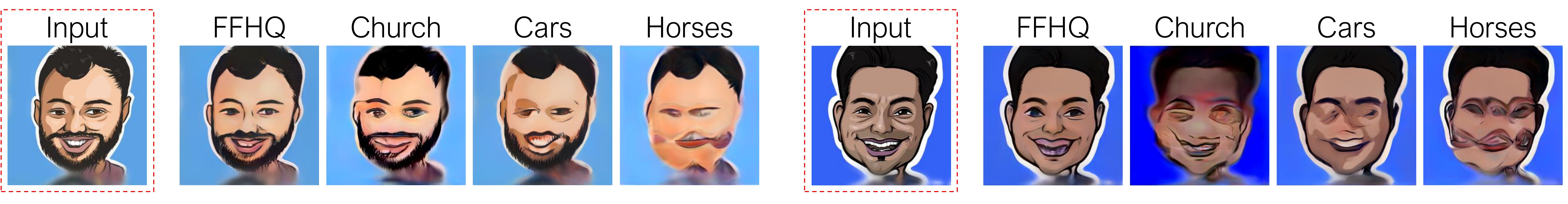}
    \caption{Embedding unseen caricature images into models adapted from different source to the same target domain (caricature). We observe that FFHQ $\rightarrow$ caricatures best captures the caricature properties, resulting in best reconstructions.}
    \label{fig:embed_full}
    \vspace{-1.5em}
\end{figure*}

\begin{figure*}[t]
    \centering
    \includegraphics[width=1.\textwidth]{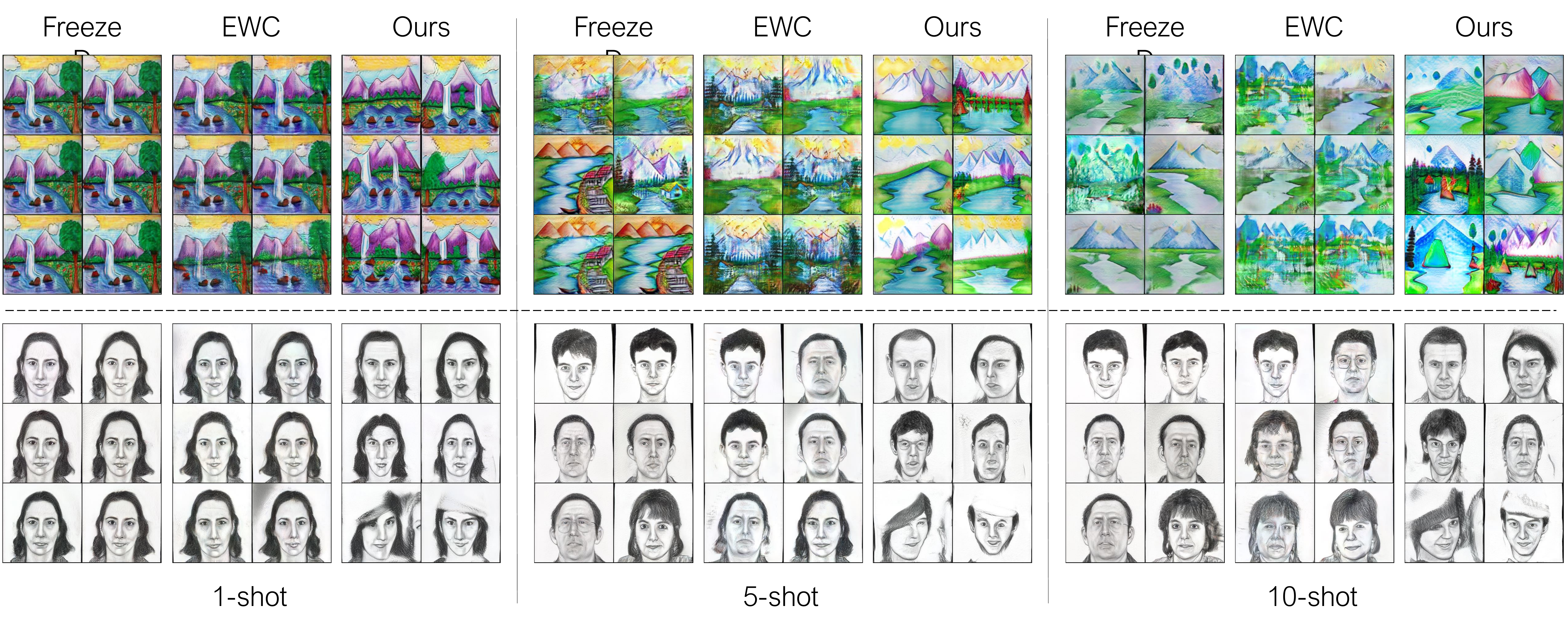}
    \caption{Comparison of our method compared to EWC~\cite{li2020fewshot} and FreezeD~\cite{mo2020freezeD} in 1-shot, 5-shot and 10-shot setting.}
    \label{fig:size_ablation_full}
    \vspace{-1.5em}
\end{figure*}

\begin{figure*}[h]
    \centering
    \includegraphics[width=1.\textwidth]{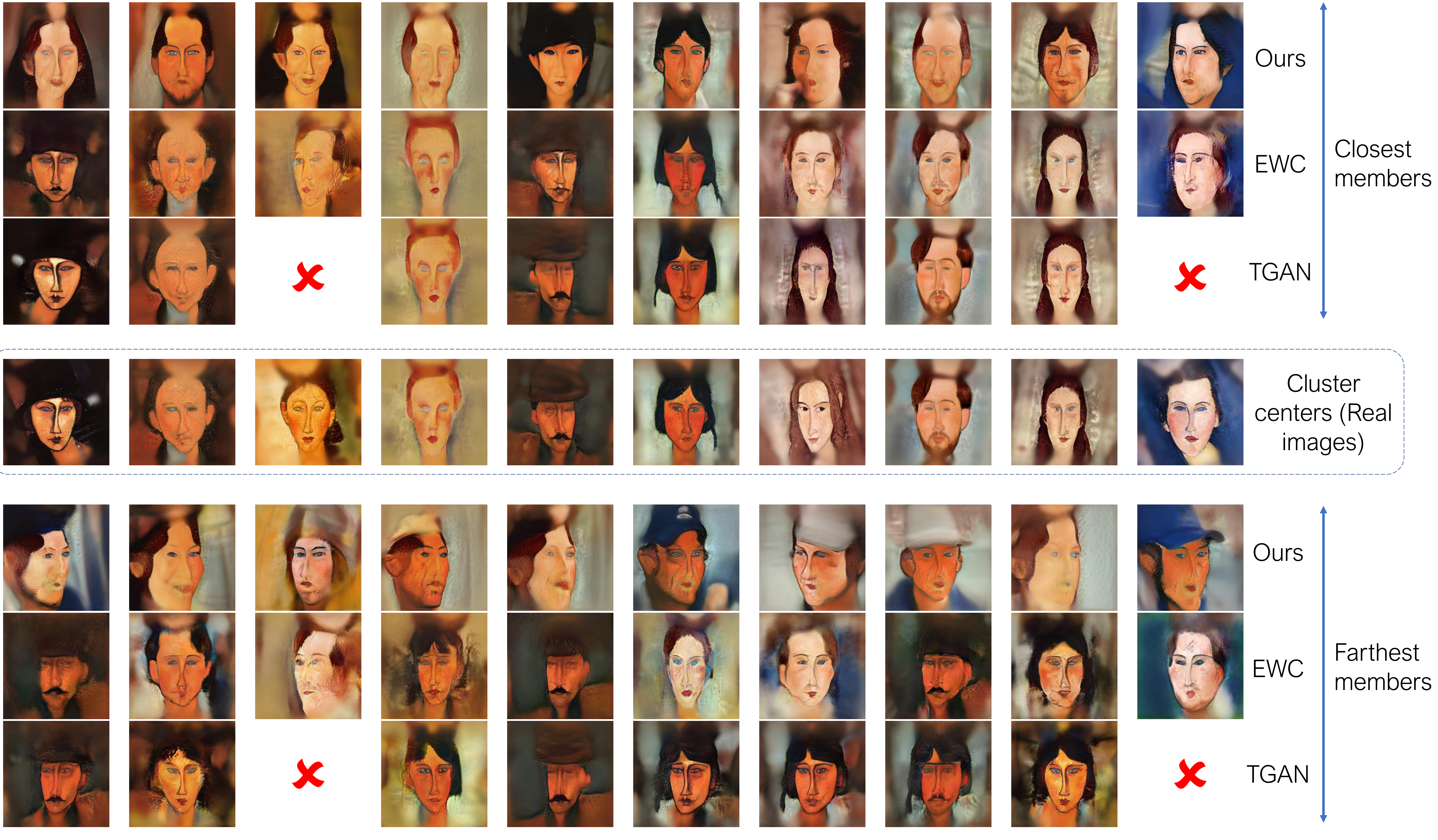}
    \caption{Visualizing the clusters formed using the technique described in Sec. 4.1 of the main paper. The closest members produced by TGAN/EWC are much more similar to the corresponding cluster center than our method, indicating that the generations using the proposed method possess more diversity.}
    \label{fig:clusters}
    \vspace{-1.5em}
\end{figure*}